\documentclass[
twocolumn,
]{ceurart}

\usepackage{listings}
\usepackage{enumitem}
\usepackage{float}
\begin{document}

\copyrightyear{2022}
\copyrightclause{Copyright for this paper by its authors.
  Use permitted under Creative Commons License Attribution 4.0
  International (CC BY 4.0).}

\conference{COMHUM 2022}

\title{Data Augmentation for Robust Character Detection in Fantasy Novels}

\author[1]{Arthur Amalvy}[orcid=0000-0002-0877-7063, email=arthur.amalvy@univ-avignon.fr]
\author[1]{Vincent Labatut}[orcid=0000-0002-2619-2835, email=vincent.labatut@univ-avignon.fr]
\author[2]{Richard Dufour}[orcid=0000-0003-1203-9108, email=richard.dufour@univ-nantes.fr]
\address[1]{Laboratoire d'Informatique d'Avignon (LIA), Avignon University, France}
\address[2]{Laboratoire des Sciences du Numérique de Nantes (LS2N), Nantes University, France}

\begin{abstract}
Named Entity Recognition (NER) is a low-level task often used as a foundation for solving higher level NLP problems. In the context of character detection in novels, NER false negatives can be an issue as they possibly imply missing certain characters or relationships completely. In this article, we demonstrate that applying a straightforward data augmentation technique allows training a model achieving higher recall, at the cost of a certain amount of precision regarding ambiguous entities. We show that this decrease in precision can be mitigated by giving the model more local context, which resolves some of the ambiguities.
\end{abstract}


\maketitle

\section{Introduction}

The \textit{Character Detection} task is concerned with detecting which characters appear in a text, and where. We decompose it in two subtasks:
\begin{itemize}
    \item Named Entity Recognition (NER), whose goal is to find character occurrences in a text.
    \item Named Entity Disambiguation (NED), which maps character occurrences to their respective normalized character form.
\end{itemize}

The result of the character detection task can be used to solve other higher-level problems. Since its output can affect the performance of other tasks relying on it, such as character network extraction~\citep{labatut-2019}, its degree of success is of importance.

\citet{dekker-2019-evaluation_ner_social_networks_novels} perform a study where they evaluate several NER models for the purpose of extracting character networks from literary texts. According to their results, performance varies greatly across novels, ranging from great to pretty low. These authors find that quite a few of the false negatives produced by the evaluated models stem from two specific types of names: word names (names that are also regular words, such as \texttt{Valor} or \texttt{Mercy}) and apostrophed names (such as \texttt{Rand al'Thor}). Replacing them with regular names allows for better detection, showing that the issue regarding these names is linked to their form and not to their surrounding context.

We suspect that these shortcomings come from the datasets used to train the NER systems assessed by \citet{dekker-2019-evaluation_ner_social_networks_novels}. Indeed, NER systems are usually trained on the main publicly available datasets (such as Ontonotes~\citep{weischedel-2011-ontonotesv5}, CoNLL-2003~\citep{tjong-2003-conll_2003_ner} or WikiGold~\citep{balasuriya-2009-wikigold}), which come from a few domains (news, web...). Among these mainstream datasets, none contain texts of the literary domain\footnote{While the Litbank dataset~\citep{bamman-2019-litbank} specifically contains only literary texts, its annotations concern \textit{nested} NER, an arguably harder task.}, and annotating new datasets is costly. This means applying off-the-shelf NER systems to literary texts suffers from these systems integrating knowledge specific only to their training data.

More precisely, there may be a mismatch in style between typical person names from the training dataset and from literary texts. This is particularly the case in the fantasy genre, where character names can have very unusual styles depending on the setting.

In order to try and fix this issue, we showcase the application of a specific data augmentation technique, \textit{mention replacement}~\citep{dai-2020-data_aug_ner}, to inject new character names in the training dataset. Using this technique, we demonstrate performance improvements for the NER task on a dataset of texts from the fantasy genre. We release our code and data, that can be used to reproduce our results, under a free license\footnote{\url{https://github.com/CompNet/ddaugNER}}.

The rest of this article is organized as follows: first, Section~\ref{sec:dataaug} provides a few examples of related data augmentation works. Then, in Section~\ref{sec:method}, we present mention replacement and the experiments we perform to assess its ability to resolve the name mismatch issue we highlighted. We discuss the results of these experiments in Section~\ref{sec:results}, and perform further analysis to shed light on the precision decrease we observe when using mention replacement. Finally, we conclude with some perspectives regarding NER performance in literary texts.

\section{Data Augmentation and Cross-Domain NER}
\label{sec:dataaug}

Generally speaking, data augmentation is the generation of new synthetic training examples. Data augmentation techniques can be used to address data scarcity problems, or to increase the diversity of the training dataset to reduce overfitting. While these techniques are ubiquitous in image processing~\citep{shorten-2019-image_da_survey}, they are less commonly explored in natural language processing~\citep{feng-2021-da_survey}, and even less for NER~\citep{dai-2020-data_aug_ner}.

A few techniques have been used to try and fix the domain discrepancy between training and testing datasets. \citet{ding-2020-daga} train a language model to generate new examples by directly including NER tags in the generated text. \citet{chen-2021-da_cross_domain_ner} train a neural architecture to transform examples from the training domain to examples closer to the test domain. Very recently, \citet{yang-2022-factmix} propose \textit{FactMix}, a two-step data augmentation process performing mention replacement followed by masked token replacement. However, as far as we know, previous works are not specifically concerned with literary texts or the mismatch in person names between domains.

\section{Method}
\label{sec:method}

In this section, we first describe mention replacement and how we apply it to texts of the fantasy genre. We then introduce the datasets we use to assess the performance of this data augmentation technique. Finally, we describe the specific setting of the experiment we carry to evaluate its performance, including the model we used and its training parameters.

\subsection{Mention Replacement}

\begin{figure*}
  \centering
  \includegraphics[scale=0.5]{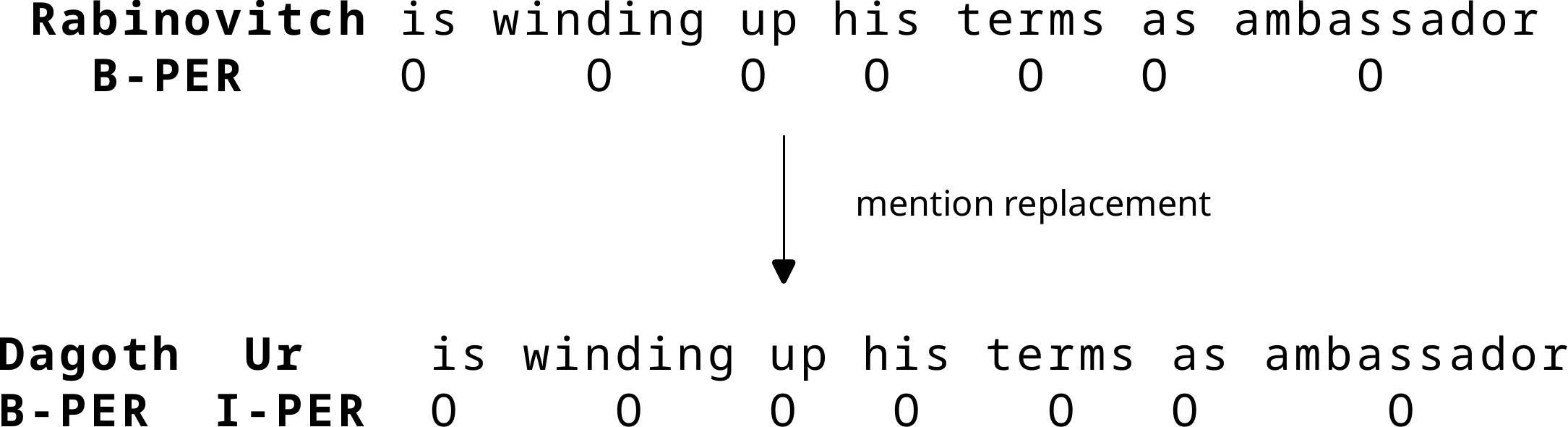}
  \caption{Example of mention replacement. Note that in this case, the sentence labels have to be slightly modified to account for the replacement entity higher number of tokens.}
  \label{fig:mention_replacement}
\end{figure*}

\citet{dai-2020-data_aug_ner} survey a few simple data augmentation schemes for NER, including mention replacement. It consists in generating new examples by replacing tokens from an entity mention found in the training dataset by tokens from another entity of the same type. Figure~\ref{fig:main_results} shows an example of this process. Based on the same principle, we propose to randomly replace training entities with ones from a list of fantasy names to increase the training dataset coverage of this type of names. If the same entity mention is present multiple times in a training example, we replace every occurrence with the same fantasy mention when performing augmentation to avoid inconsistencies. We hope that injecting typical fantasy names into the training dataset using mention replacement will allow the model to better detect them.

To perform mention replacement, we compose a list of replacement entities that do not come directly from the evaluation dataset. We scrap the entirety of the names from \textit{The Elder Scrolls} series of video games that are mentioned on the \textit{Unofficial Elder Scrolls Pages}, a wiki dedicated to \textit{The Elder Scrolls} universe\footnote{\url{https://en.uesp.net/wiki/Main_Page}}. In total, we retrieve 22{,}748 first names, 4{,}879 last names, 647 suffixes and 37 prefixes. We combine these to form new mentions: when performing mention replacement, we randomly compose a new mention by first sampling from a set of valid forms (\texttt{[first name]+[last name]}, \texttt{[prefix]+[first name]+[last name]}, ...). We weight this sampling by a rough approximation of each form's frequency. When a form has been selected, we uniformly sample a name part for each of the form's elements. 

Since we are interested in knowing the impact of the number of generated examples, we compare the performance of models trained with different augmentation rates. We define the \textit{augmentation rate} as the ratio of examples generated over the dataset size, so an augmentation rate of 1 would mean generating as many examples as there are examples in the dataset.

\subsection{Datasets}

\paragraph{Training Dataset} We train our model on a modified version of CoNLL-2003~\citep{tjong-2003-conll_2003_ner}, which is one of the best known and used NER dataset. CoNLL-2003 is composed of 14{,}041 sentences from news articles. Our modifications consist in including honorifics as part of entities to be consistent with our evaluation dataset (see below). The training dataset contains annotations for four classes: persons (\texttt{PER}), organizations (\texttt{ORG}), locations (\texttt{LOC}) and miscellaneous (\texttt{MISC}). Despite restricting ourselves to character detection (\texttt{PER} class), we keep annotations for every class since we observe that training the model with all classes increase the performance of NER.

\paragraph{Evaluation dataset} We use a corrected version of the subset of fantasy novels from the dataset of~\citet{dekker-2019-evaluation_ner_social_networks_novels}, where only persons entities (\texttt{PER}) are annotated. We found the dataset had to be corrected because we noticed a number of encoding, tokenization and annotation issues. We fixed the encoding and tokenization issues manually. In order to consistently correct annotation errors, we designed an annotation guide and applied a semi-automated correction process. It consists of 3 steps:

\begin{enumerate}[leftmargin=\parindent] 
\item We apply a number of simple heuristics to identify obvious errors:
\begin{itemize}[leftmargin=\parindent]
    \item When a span is not annotated as a person occurrence, but it appears in the list of character names for the book, it might be a false negative.
    \item When a span is annotated as a person occurrence, but it does not appear in this list, it might a false positive.
    \item When a span is annotated as a person occurrence, but the first letter of all of its tokens is not capitalized, it might be a false positive.
\end{itemize}
Each time one of these heuristics recommends an annotation change, we manually check if it is correct before accepting it.
\item We then consider the differences between the existing annotations and the predictions of a BERT model~\citep{devlin-2019-bert} fine-tuned for the NER task, and manually reconcile them.
\item Finally, we manually correct the few remaining errors that we could detect.
\end{enumerate}

\begin{table}
  \centering
  \begin{tabular}{lrrr}
                                & Exact   & Partial &         \\
    Name Set                    & Match   & Match   & Unseen  \\
    \hline
    \texttt{CoNLL-2003} (train) &  9.62\% & 76.97\% & 13.41\% \\
    \texttt{The Elder Scrolls}  & 12.25\% & 76.38\% & 11.37\%
  \end{tabular}
  \caption{Overlap of different name sets with the set of character names from our evaluation dataset. The "exact match" column reports the proportion of \texttt{PER} labeled tokens from our evaluation dataset that are also in the studied name set. The "partial match" column is the proportion of \texttt{PER} labeled tokens containing a wordpiece that exists in the studied name set. Finally, the "unseen" column indicates the rest of \texttt{PER} labeled tokens from the evaluation dataset.}
  \label{tab:overlap}
\end{table}

The dataset consists in the first chapter of 17 novels. Together, these chapters are composed of 5{,}518 sentences and contain 360 unique person names. Inspired by \citet{taille-2020-ner_lexicality}, we report in Table~\ref{tab:overlap} the name overlap between our evaluation dataset and:
\begin{itemize}
    \item The train portion of the CoNLL-2003~\citep{tjong-2003-conll_2003_ner};
    \item Our list of names from \textit{The Elder Scrolls}, used for mention replacement.
\end{itemize}

\subsection{Experiment}

We fine-tune a BERT model~\citep{devlin-2019-bert} with an added classification head on our training dataset, and report its performance on our evaluation dataset depending on the augmentation rate. In order to obtain stable results, we report the mean of the results from 10 fine-tuning runs for each augmentation rate that we test. Models without augmentations are trained for two full epochs, while we adjust the number of training steps for model with augmentation so that they learn on the same number of examples as models without augmentation.
We use a learning rate of $2 \cdot 10^{-5}$~\citep{devlin-2019-bert}, and we initialize each model with the \texttt{bert-base-cased} checkpoint of the \textit{transformers} library~\citep{wolf-2020-transformers}. At inference time, each model predicts NER tags separately for each sentence, and we do not provide it with any additional context. Since the evaluation dataset only contains annotations for the \texttt{PER} class, we simply discard predictions made for other classes.

\section{Results}
\label{sec:results}

\begin{table}
  \centering
  \begin{tabular}{|c|c|c|c|c|}
    \hline  
    Vin      & Camon   & Bilbo & Pug    & Arlen    \\
    \hline
    Galladon & Bran    & Selia & Bug    & Cob      \\
    \hline
    Raoden   & Gandalf & Szeth & Weasel & Chivalry \\
    \hline
    Logen    & Theron  & Frodo & Jory   & Jezrien  \\
    \hline
  \end{tabular}
  \caption{Top 20 character names that were better detected by a model trained with data augmentation (10\% augmentation rate) but not by a model trained with the original dataset only.}
  \label{tab:recalled_names}
\end{table}

\begin{figure}[h!]
  \centering
  \includegraphics[width=\linewidth]{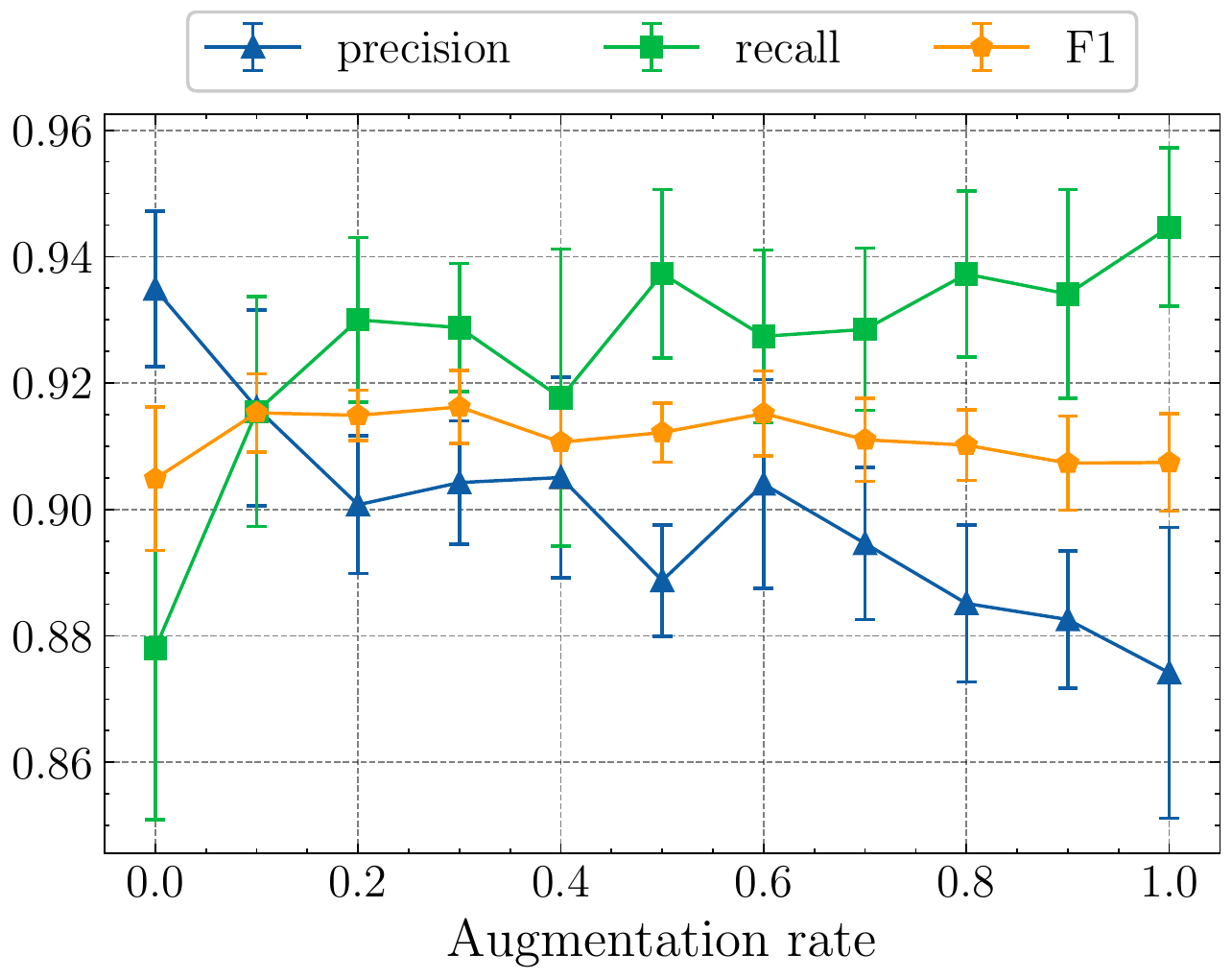}
  \caption{Precision, recall and F1-score against augmentation rate. Error bars represent the 95\% confidence interval.}
  \label{fig:main_results}
\end{figure}

Results can be seen in Figure~\ref{fig:main_results}. We report performance according to the CoNLL-2003 guidelines~\citep{tjong-2003-conll_2003_ner} and using the Python \texttt{seqeval} library~\citep{nakayama-2018-seqeval}. We observe a notable increase in recall using augmentation, while precision decreases. The recall increase can be attributed to the model picking up on more unusual character names, as Table~\ref{tab:recalled_names} shows.

To try to understand the decrease in precision when using data augmentation, we perform some additional experiments. Based on our manual investigation of the false positives, we formulate two hypotheses that we want to test:

\begin{itemize}
    \item[$h_1$] Adding more examples with \texttt{PER} entities modifies the class distribution in the dataset, progressively leading to a situation of class imbalance where these entities are way more frequent than those of the other classes (\texttt{LOC}, \texttt{ORG} and \texttt{MISC}). As we observe that removing other entity classes from the training dataset is detrimental to \texttt{PER} detection performance (possibly because having more classes means the model is better at distinguishing between them), we suppose that class imbalance can be detrimental as well.
    \item[$h_2$] Increasing name variety in the training set encourages the model to try and make \texttt{PER} predictions for tokens whose classes are ambiguous given a specific sentence. We deem a token ``ambiguous'' when the context given to the model does not suffice to infer its correct class (for example, when it is not possible to distinguish between a \texttt{PER} and a \texttt{LOC} entity using only an input sentence).
\end{itemize}

\subsection{Class Imbalance}

\begin{figure*}
  \centering
  \includegraphics[width=\textwidth]{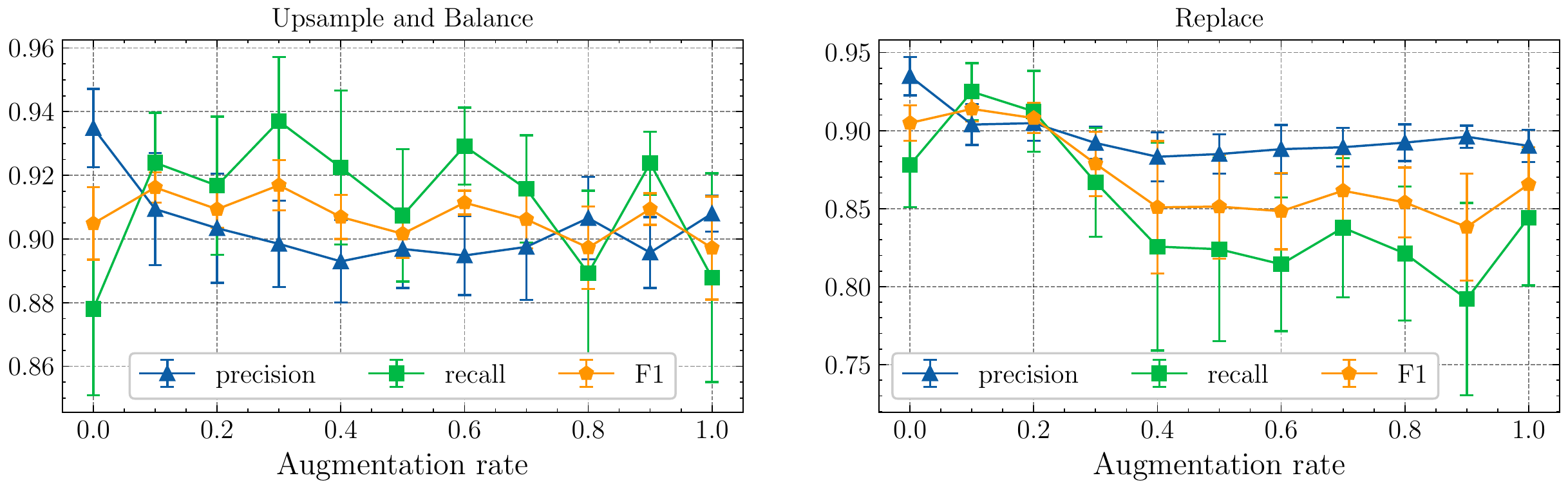}
  \caption{Precision, recall and F1-score against augmentation rate for different augmentation methods. Error bars represent the 95\% confidence interval.}
  \label{fig:aug_methods}
\end{figure*}

To check if the decrease in precision is due to class imbalance, we test two different methods to inject names in the training dataset that should not suffer from the class imbalance problem:

\begin{itemize}
\item \texttt{Upsample and balance}: After adding newly generated examples to the dataset, we copy some of the original examples and add them to the dataset to restore the original class distribution.
\item \texttt{Replace}: Instead of \textit{adding} newly generated examples to the dataset, we \textit{substitute} them for the original training samples.
\end{itemize}

Figure~\ref{fig:aug_methods} shows the performance of a model trained on a dataset modified with the above augmentation methods. The \textit{upsample and balance} strategy still increases recall, but does not fix the precision issue. Meanwhile, the \texttt{replace} strategy still makes precision decrease, and also generally decreases performance for an augmentation rate greater than 20\%.

As none of these two methods are able to fix the precision issue, we conclude that class imbalance is not a plausible cause of precision decreasing.

\subsection{Name Variety and Ambiguous Occurrences}

In order to check $h_2$, we analyze the difference in false positives between a model trained with augmentation and a model trained without it. As errors can vary between runs, we perform three training runs for each model and keep the false positives that are consistent between the runs. We then observe the set of false positives of the model trained with augmentation that are not present in the set of false positives of the model trained without augmentation. This allows us to analyze errors that are specific to our augmentation scheme.

We manually find that for 62\% (58/93) of these false positives, the model attributes a \texttt{PER} label to some tokens even though the context alone is not sufficient to know the correct label of the entity, such as in the following example from \textit{Elantris}:

\smallskip
\fbox{\parbox{0.9\columnwidth}{\texttt{Raoden stood, and as he did, his eyes fell on Elantris again.}}}
\smallskip

\noindent Here, the model correctly predicts that \texttt{Elantris} is a named entity, but gives it the class \texttt{PER} while \texttt{Elantris} is actually a city (which corresponds to the \texttt{LOC} class). Of course, predicting the correct class of \texttt{Elantris} in this example would only be a matter of luck even for a perfect model, as the given context alone is not sufficient.

Overall, this result shows a bias towards the \texttt{PER} class, that was introduced by increasing the variety of possible names at training time.

To try and mitigate this effect, we supply the model with broader context, in hope that this context can help resolving some ambiguities. In our previous example, the next sentence gives enough context to predict the correct entity type for \texttt{Elantris}:

\smallskip
\fbox{\parbox{0.9\columnwidth}{\texttt{Raoden stood, and as he did, his eyes fell on Elantris again. Resting in the great city's shadow, Kae seemed like an insignificant village by comparison.}}}
\smallskip

\noindent We define a context size of $n$ as giving the model the $n$ sentences that \textit{precede} the considered sentence, as well as the $n$ sentences that \textit{follow} it.

\begin{figure*}
  \centering
  \includegraphics[width=\textwidth]{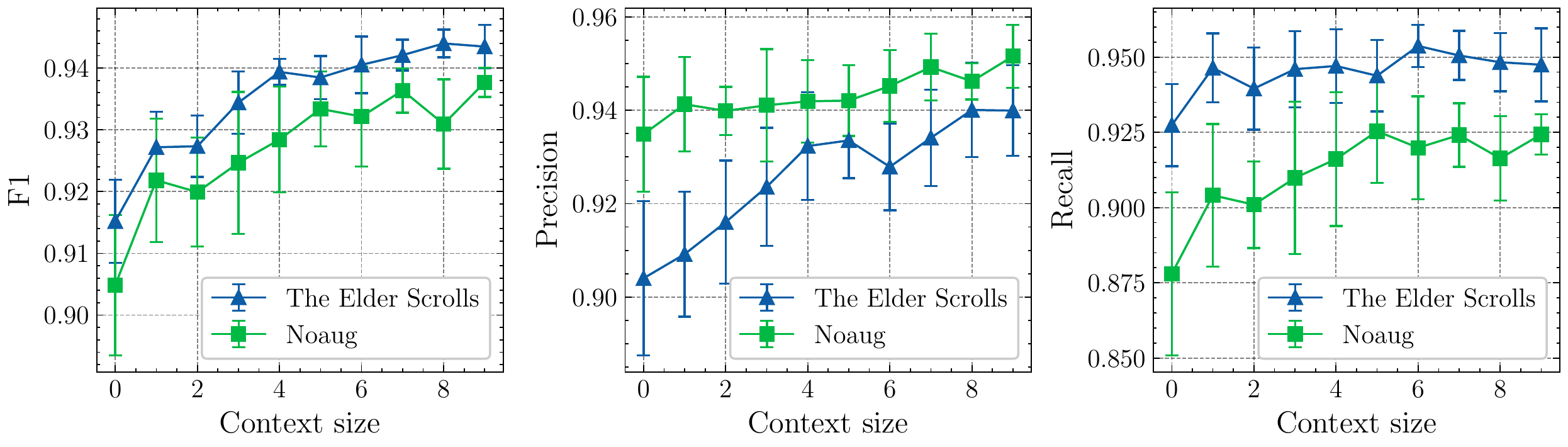}
  \caption{F1-score, precision and recall against different context sizes for a model trained without augmentation (\texttt{Noaug}) and a model trained with augmentation (\texttt{The Elder Scrolls}) (60\% augmentation ratio). Error bars representent the 95\% confidence interval.}
  \label{fig:context}
\end{figure*}

Figure~\ref{fig:context} shows the effect of giving more context to models trained with augmentation (\texttt{The Elder Scrolls}) or without it (\texttt{Noaug}). As can be seen, context is beneficial to both models, steadily increasing F1-score.  Precision increases a lot for the model trained with augmentation, as previously ambiguous entities are now correctly labeled. Using our previously described method for error analysis, we find that only 48\% (30/63) of the entities found in false positives are deemed ambiguous for a context size of 1. Meanwhile, the recall augmentation for the \texttt{Noaug} model can be attributed to it predicting \texttt{PER} entities in previously ambiguous cases. While adding more context results in greater recall for the \texttt{Noaug} model, it still never surpasses the recall of the \texttt{The Elder Scrolls} model.

\section{Conclusion}

We demonstrated the usage of a simple data augmentation technique to better detect fantasy names when performing named entity recognition. While this technique greatly increases model recall, precision decreases as the model sometimes does not have enough information to predict the correct class of some entities. We showed that this issue can be mitigated by giving the model more local information. However, for the same context size, our technique always result in greater recall but lower precision. In the context of character detection, we argue that this trade is beneficial: while false positives can be filtered after performing NER (automatically or manually), false negatives are not easily recovered. We also note that increasing NER context has a positive impact on recall in general.

Further work might investigate the role of context in NER. While we showed that it can be used to disambiguate some mentions, we were only interested in local context. Nevertheless, necessary disambiguation information is not always present in the local context, And might exist in other places for each novel.

\bibliography{biblio}

\begin{thebibliography}{16}
\expandafter\ifx\csname natexlab\endcsname\relax\def\natexlab#1{#1}\fi
\providecommand{\url}[1]{\texttt{#1}}
\providecommand{\href}[2]{#2}
\providecommand{\path}[1]{#1}
\providecommand{\DOIprefix}{doi:}
\providecommand{\ArXivprefix}{arXiv:}
\providecommand{\URLprefix}{URL: }
\providecommand{\Pubmedprefix}{pmid:}
\providecommand{\doi}[1]{\href{http://dx.doi.org/#1}{\path{#1}}}
\providecommand{\Pubmed}[1]{\href{pmid:#1}{\path{#1}}}
\providecommand{\bibinfo}[2]{#2}
\ifx\xfnm\relax \def\xfnm[#1]{\unskip,\space#1}\fi
\bibitem[{Labatut and Bost(2019)}]{labatut-2019}
\bibinfo{author}{V.~Labatut}, \bibinfo{author}{X.~Bost},
\newblock \bibinfo{title}{Extraction and analysis of fictional character
  networks : A survey},
\newblock \bibinfo{journal}{ACM Computing Surveys} \bibinfo{volume}{52}
  (\bibinfo{year}{2019}) \bibinfo{pages}{89}. \DOIprefix\doi{10.1145/3344548}.
\bibitem[{Dekker et~al.(2019)Dekker, Kuhn, and van
  Erp}]{dekker-2019-evaluation_ner_social_networks_novels}
\bibinfo{author}{N.~Dekker}, \bibinfo{author}{T.~Kuhn}, \bibinfo{author}{M.~van
  Erp},
\newblock \bibinfo{title}{Evaluating named entity recognition tools for
  extracting social networks from novels},
\newblock \bibinfo{journal}{PeerJ Computer Science} \bibinfo{volume}{5}
  (\bibinfo{year}{2019}) \bibinfo{pages}{e189}.
  \DOIprefix\doi{10.7717/peerj-cs.189}.
\bibitem[{Weischedel et~al.(2011)Weischedel, Hovy, Marcus, Palmer, Belvin,
  Pradhan, Ramshaw, and Xue}]{weischedel-2011-ontonotesv5}
\bibinfo{author}{R.~Weischedel}, \bibinfo{author}{E.~Hovy},
  \bibinfo{author}{M.~Marcus}, \bibinfo{author}{M.~Palmer},
  \bibinfo{author}{R.~Belvin}, \bibinfo{author}{S.~Pradhan},
  \bibinfo{author}{L.~Ramshaw}, \bibinfo{author}{N.~Xue},
  \bibinfo{title}{{OntoNotes}: A large training corpus for enhanced
  processing}, \bibinfo{year}{2011}. \URLprefix
  \url{https://www.cs.cmu.edu/~hovy/papers/09OntoNotes-GALEbook.pdf}.
\bibitem[{Tjong Kim~Sang and De~Meulder(2003)}]{tjong-2003-conll_2003_ner}
\bibinfo{author}{E.~F. Tjong Kim~Sang}, \bibinfo{author}{F.~De~Meulder},
\newblock \bibinfo{title}{Introduction to the {C}o{NLL}-2003 shared task:
  Language-independent named entity recognition},
\newblock in: \bibinfo{booktitle}{7th Conference on Natural Language Learning},
  \bibinfo{year}{2003}, pp. \bibinfo{pages}{142--147}. \URLprefix
  \url{https://aclanthology.org/W03-0419}.
\bibitem[{Balasuriya et~al.(2009)Balasuriya, Ringland, Nothman, Murphy, and
  Curran}]{balasuriya-2009-wikigold}
\bibinfo{author}{D.~Balasuriya}, \bibinfo{author}{N.~Ringland},
  \bibinfo{author}{J.~Nothman}, \bibinfo{author}{T.~Murphy},
  \bibinfo{author}{J.~R. Curran},
\newblock \bibinfo{title}{Named entity recognition in {W}ikipedia},
\newblock in: \bibinfo{booktitle}{Workshop on The People's Web Meets {NLP}:
  Collaboratively Constructed Semantic Resources}, \bibinfo{year}{2009}, pp.
  \bibinfo{pages}{10--18}. \URLprefix \url{https://aclanthology.org/W09-3302}.
\bibitem[{Bamman et~al.(2019)Bamman, Popat, and Shen}]{bamman-2019-litbank}
\bibinfo{author}{D.~Bamman}, \bibinfo{author}{S.~Popat},
  \bibinfo{author}{S.~Shen},
\newblock \bibinfo{title}{An annotated dataset of literary entities},
\newblock in: \bibinfo{booktitle}{Conference of the North {A}merican Chapter of
  the Association for Computational Linguistics: Human Language Technologies},
  volume~\bibinfo{volume}{1}, \bibinfo{year}{2019}, pp.
  \bibinfo{pages}{2138--2144}. \DOIprefix\doi{10.18653/v1/N19-1220}.
\bibitem[{Dai and Adel(2020)}]{dai-2020-data_aug_ner}
\bibinfo{author}{X.~Dai}, \bibinfo{author}{H.~Adel},
\newblock \bibinfo{title}{An analysis of simple data augmentation for named
  entity recognition},
\newblock in: \bibinfo{booktitle}{International Conference on Computational
  Linguistics}, \bibinfo{year}{2020}, pp. \bibinfo{pages}{3861--3867}.
  \DOIprefix\doi{10.18653/v1/2020.coling-main.343}.
\bibitem[{Shorten and Khoshgoftaar(2019)}]{shorten-2019-image_da_survey}
\bibinfo{author}{C.~Shorten}, \bibinfo{author}{T.~M. Khoshgoftaar},
\newblock \bibinfo{title}{A survey on image data augmentation for deep
  learning},
\newblock \bibinfo{journal}{Journal of Big Data} \bibinfo{volume}{6}
  (\bibinfo{year}{2019}) \bibinfo{pages}{60}.
  \DOIprefix\doi{10.1186/s40537-019-0197-0}.
\bibitem[{Feng et~al.(2021)Feng, Gangal, Wei, Chandar, Vosoughi, Mitamura, and
  Hovy}]{feng-2021-da_survey}
\bibinfo{author}{S.~Y. Feng}, \bibinfo{author}{V.~Gangal},
  \bibinfo{author}{J.~Wei}, \bibinfo{author}{S.~Chandar},
  \bibinfo{author}{S.~Vosoughi}, \bibinfo{author}{T.~Mitamura},
  \bibinfo{author}{E.~Hovy},
\newblock \bibinfo{title}{A survey of data augmentation approaches for {NLP}},
\newblock in: \bibinfo{booktitle}{Findings of the Association for Computational
  Linguistics: ACL-IJCNLP 2021}, \bibinfo{year}{2021}, pp.
  \bibinfo{pages}{968--988}. \DOIprefix\doi{10.18653/v1/2021.findings-acl.84}.
\bibitem[{Ding et~al.(2020)Ding, Liu, Bing, Kruengkrai, Nguyen, Joty, Si, and
  Miao}]{ding-2020-daga}
\bibinfo{author}{B.~Ding}, \bibinfo{author}{L.~Liu}, \bibinfo{author}{L.~Bing},
  \bibinfo{author}{C.~Kruengkrai}, \bibinfo{author}{T.~H. Nguyen},
  \bibinfo{author}{S.~Joty}, \bibinfo{author}{L.~Si},
  \bibinfo{author}{C.~Miao},
\newblock \bibinfo{title}{{DAGA}: Data augmentation with a generation approach
  for low-resource tagging tasks},
\newblock in: \bibinfo{booktitle}{Conference on Empirical Methods in Natural
  Language Processing}, \bibinfo{year}{2020}, pp. \bibinfo{pages}{6045--6057}.
  \DOIprefix\doi{10.18653/v1/2020.emnlp-main.488}.
\bibitem[{Chen et~al.(2021)Chen, Aguilar, Neves, and
  Solorio}]{chen-2021-da_cross_domain_ner}
\bibinfo{author}{S.~Chen}, \bibinfo{author}{G.~Aguilar},
  \bibinfo{author}{L.~Neves}, \bibinfo{author}{T.~Solorio},
\newblock \bibinfo{title}{Data augmentation for cross-domain named entity
  recognition},
\newblock in: \bibinfo{booktitle}{Conference on Empirical Methods in Natural
  Language Processing}, \bibinfo{year}{2021}, pp. \bibinfo{pages}{5346--5356}.
  \DOIprefix\doi{10.18653/v1/2021.emnlp-main.434}.
\bibitem[{Yang et~al.(2022)Yang, Yuan, Cui, Gao, and Zhang}]{yang-2022-factmix}
\bibinfo{author}{L.~Yang}, \bibinfo{author}{L.~Yuan}, \bibinfo{author}{L.~Cui},
  \bibinfo{author}{W.~Gao}, \bibinfo{author}{Y.~Zhang},
\newblock \bibinfo{title}{Factmix: Using a few labeled in-domain examples to
  generalize to cross-domain named entity recognition},
\newblock \bibinfo{journal}{arXiv} \bibinfo{volume}{cs.CL}
  (\bibinfo{year}{2022}) \bibinfo{pages}{2208.11464}.
  \DOIprefix\doi{10.48550/ARXIV.2208.11464}.
\bibitem[{Devlin et~al.(2019)Devlin, Chang, Lee, and
  Toutanova}]{devlin-2019-bert}
\bibinfo{author}{J.~Devlin}, \bibinfo{author}{M.~Chang},
  \bibinfo{author}{K.~Lee}, \bibinfo{author}{K.~Toutanova},
\newblock \bibinfo{title}{{BERT}: Pre-training of deep bidirectional
  transformers for language understanding},
\newblock in: \bibinfo{booktitle}{Conference of the North {A}merican Chapter of
  the Association for Computational Linguistics: Human Language Technologies},
  volume~\bibinfo{volume}{1}, \bibinfo{year}{2019}, pp.
  \bibinfo{pages}{4171--4186}. \DOIprefix\doi{10.18653/v1/N19-1423}.
\bibitem[{Taillé et~al.(2020)Taillé, Guigue, and
  Gallinari}]{taille-2020-ner_lexicality}
\bibinfo{author}{B.~Taillé}, \bibinfo{author}{V.~Guigue},
  \bibinfo{author}{P.~Gallinari},
\newblock \bibinfo{title}{Contextualized embeddings in named-entity
  recognition: An empirical study on generalization},
\newblock in: \bibinfo{booktitle}{Advances in Information Retrieval},
  \bibinfo{year}{2020}, pp. \bibinfo{pages}{383--391}.
  \DOIprefix\doi{10.1007/978-3-030-45442-5_48}.
\bibitem[{Wolf et~al.(2020)Wolf, Debut, Sanh, Chaumond, Delangue, Moi, Cistac,
  Rault, Louf, Funtowicz, Davison, Shleifer, von Platen, Ma, Jernite, Plu, Xu,
  Le~Scao, Gugger, Drame, Lhoest, and Rush}]{wolf-2020-transformers}
\bibinfo{author}{T.~Wolf}, \bibinfo{author}{L.~Debut},
  \bibinfo{author}{V.~Sanh}, \bibinfo{author}{J.~Chaumond},
  \bibinfo{author}{C.~Delangue}, \bibinfo{author}{A.~Moi},
  \bibinfo{author}{P.~Cistac}, \bibinfo{author}{T.~Rault},
  \bibinfo{author}{R.~Louf}, \bibinfo{author}{M.~Funtowicz},
  \bibinfo{author}{J.~Davison}, \bibinfo{author}{S.~Shleifer},
  \bibinfo{author}{P.~von Platen}, \bibinfo{author}{C.~Ma},
  \bibinfo{author}{Y.~Jernite}, \bibinfo{author}{J.~Plu},
  \bibinfo{author}{C.~Xu}, \bibinfo{author}{T.~Le~Scao},
  \bibinfo{author}{S.~Gugger}, \bibinfo{author}{M.~Drame},
  \bibinfo{author}{Q.~Lhoest}, \bibinfo{author}{A.~M. Rush},
\newblock \bibinfo{title}{Transformers: State-of-the-art natural language
  processing},
\newblock in: \bibinfo{booktitle}{Conference on Empirical Methods in Natural
  Language Processing: System Demonstrations}, \bibinfo{year}{2020}, pp.
  \bibinfo{pages}{38--45}. \URLprefix
  \url{https://www.aclweb.org/anthology/2020.emnlp-demos.6}.
\bibitem[{Nakayama(2018)}]{nakayama-2018-seqeval}
\bibinfo{author}{H.~Nakayama}, \bibinfo{title}{{seqeval}: A python framework
  for sequence labeling evaluation}, \bibinfo{year}{2018}. \URLprefix
  \url{https://github.com/chakki-works/seqeval}.

\end{thebibliography}

\end{document}